\documentclass[runningheads]{llncs}
\usepackage[T1]{fontenc}
\usepackage{graphicx}

\usepackage{multirow}
\usepackage{amsmath}
\usepackage{amssymb}
\usepackage{subcaption}
\usepackage{wrapfig}
\usepackage{algorithm}
\usepackage{algpseudocode}
\usepackage{pifont}
\usepackage{siunitx}
\usepackage{makecell}
\usepackage[utf8]{inputenc}
\usepackage{hyperref}
\usepackage{url}
\usepackage{booktabs}
\usepackage{amsfonts}
\usepackage{nicefrac}
\usepackage{microtype}
\usepackage{xcolor}
\usepackage{enumitem}
\usepackage{soul}

\newcommand{\cmark}{\ding{51}}
\newcommand{\xmark}{\ding{55}}

\definecolor{mygray}{gray}{.9}

\newcommand{\citep}[1]{\cite{#1}}
\newcommand{\citet}[1]{\cite{#1}}

\usepackage{amsmath,amsfonts,bm}

\def\eqref#1{equation~\ref{#1}}

\def\1{\bm{1}}

\DeclareMathAlphabet{\mathsfit}{\encodingdefault}{\sfdefault}{m}{sl}
\SetMathAlphabet{\mathsfit}{bold}{\encodingdefault}{\sfdefault}{bx}{n}

\begin{document}

\title{PolypSteer: Counterfactual Endoscopic Synthesis via Training-Free Activation Steering}
\titlerunning{PolypSteer: Counterfactual Endoscopic Synthesis}

\author{Trong Thang Pham\inst{1}\thanks{Corresponding author: \email{tp030@uark.edu}.} \and
Loc Nguyen\inst{1} \and
Anh Nguyen\inst{2} \and
Hien V. Nguyen\inst{3} \and
Ngan Le\inst{1}}
\index{Pham, Trong Thang}
\index{Nguyen, Loc}
\index{Nguyen, Anh}
\index{Nguyen, Hien V.}
\index{Le, Ngan}
\authorrunning{T.T. Pham et al.}
\institute{University of Arkansas, Fayetteville AR, USA\\
\email{\{tp030,pnguyen4,thile\}@uark.edu} \and
University of Liverpool, Liverpool, UK\\
\email{anh.nguyen@liverpool.ac.uk} \and
University of Houston, Houston TX, USA\\
\email{hvnguy35@central.uh.edu}}
\maketitle

\begin{abstract}
Generative diffusion models are increasingly used for medical imaging data augmentation, but text prompting cannot produce \emph{causal} training data. Re-prompting rerolls the entire generation trajectory, altering anatomy, texture, and background. Inversion-based editing methods introduce reconstruction error that causes structural drift. We propose \textbf{PolypSteer}, a training-free activation-steering framework for endoscopic synthesis. PolypSteer identifies a pathology vector for each contrastive prompt pair in the cross-attention layers of a diffusion transformer. At inference time, it steers image activations along this vector, generating counterfactual pairs from scratch where the \emph{only} difference is the steered concept. All other structure is preserved by construction. We evaluate PolypSteer across three experiments on Kvasir v3 and HyperKvasir. On counterfactual generation across three clinical concept pairs, PolypSteer achieves flip rates of 0.800, 0.925, and 0.950, outperforming the best inversion-based baseline in both concept flip rate and structural preservation. On dye disentanglement, PolypSteer achieves 75\% dye removal against 20\% (PnP) and 10\% (h-Edit). On downstream polyp detection, augmenting with PolypSteer counterfactual pairs achieves ViT AUC of 0.9755 versus 0.9083 for quantity-matched re-prompting, confirming that counterfactual structure drives the gain. Code is available at \url{https://github.com/UARK-AICV/PolypSteer}.

\end{abstract}

\section{Introduction}
\label{sec:intro}

A key challenge in endoscopic image analysis is training pathology detectors that respond to disease-specific features rather than confounding anatomy. A natural solution is to generate anatomy-matched pairs of diseased and healthy images so the detector learns only the targeted concept. Diffusion models can in principle generate such pairs, but text-to-image re-prompting between any two concepts (``polyp'' to ``normal'', ``colitis'' to healthy mucosa, or dyed to undyed tissue) rerolls the entire generation trajectory, producing a completely different image.

Diffusion-based editing methods~\cite{hedit2024,tumanyan2023plug} start from a source image but rely on Denoising Diffusion Implicit Models (DDIM) inversion, an approximation that introduces reconstruction error. RadEdit~\cite{radedit2024} compounds this with mandatory mask annotations per region. No workaround closes this gap. Even improved inversion~\cite{mokady_null_2023} cannot eliminate drift, because re-entering a trajectory is inherently approximate. The requirement that non-targeted structure\footnote{By \emph{non-targeted} we mean all structure unrelated to the steered concept: anatomy, texture, and background that should remain identical between the pair.} be exactly preserved, not merely approximated, remains unmet.

We propose \textbf{PolypSteer}, a training-free framework requiring no fine-tuning, source images, or annotations. It estimates pathology vectors from contrastive prompt pairs in the cross-attention space of a frozen diffusion transformer and steers activations along those vectors at inference time. With a shared noise seed, both images traverse the \emph{same} trajectory, making all non-targeted concept structure identical by construction.

Our contributions are:
\begin{enumerate}[nosep,leftmargin=*]
    \item We introduce a \textbf{cosine-similarity-derived activation steering} mechanism that steers clinical concepts by removing only the concept-aligned component of each cross-attention token via a per-token gate. This gate reveals exactly where and when the model steers, providing \textbf{built-in spatial interpretability} absent from inversion-based approaches.
    \item We demonstrate \textbf{inversion-free counterfactual pair generation} from scratch. Both images share the same noise trajectory through a frozen model. PolypSteer outperforms existing state-of-the-art editing methods in non-targeted concept structure preservation.
\end{enumerate}

\section{Related Work}
\label{sec:related}

\noindent\textbf{Diffusion-Based Editing and Medical Counterfactual Generation.}
Training-based editing approaches fine-tune the denoiser or auxiliary networks~\cite{kim2022diffusionclip,kawar2023imagic,kwon2023diffusion}, while training-free methods manipulate attention maps~\cite{cao2023masactrl,parmar2023zero}, improve inversion fidelity~\cite{mokady_null_2023,huberman2024edit,brack2024ledits}, or use masks to localise edits~\cite{avrahami_blended_2022,couairon_diffedit_2022}.
In the medical domain, counterfactual and pathology-removal methods~\cite{pawlowski_dscm_2020,sanchez_healthy_2022,fontanella_diffusion_2023,gu_biomedjourney_2023} and RadEdit~\cite{radedit2024} all follow the same generate $\to$ invert $\to$ edit pipeline, inheriting the fundamental limitation that DDIM inversion is an approximation and background drift is bounded but never eliminated.
\textit{PolypSteer avoids this pipeline by generating both images from the same noise seed, guaranteeing non-targeted structure preservation.}
\noindent\textbf{Concept Control via Latent Directions.}
Recent work shows that diffusion activations encode semantically meaningful linear directions~\cite{kwon2023diffusion,DBLP:conf/nips/ParkKCJU23,DBLP:conf/cvpr/SiH0024,tumanyan2023plug} that can steer generation.
SDID~\cite{DBLP:conf/cvpr/0010S00G24} and SAeUron~\cite{cywinski2025saeuron} require extensive training, while concept-erasure methods fine-tune weights~\cite{DBLP:conf/iccv/GandikotaMFB23,DBLP:conf/iccv/KumariZWS0Z23} or Low-Rank Adaptation (LoRA) adapters~\cite{DBLP:conf/cvpr/Lyu0HCJ00HD24,DBLP:conf/cvpr/LuWLLK24}, all prohibitive when labelled medical data is scarce.
None has been applied to medical imaging, where class labels encode \emph{entangled} visual attributes that text cannot decompose, and utility must be validated on downstream clinical tasks rather than perceptual metrics alone.
\textit{PolypSteer addresses this gap in that it requires no training, operates on entangled medical class labels, and validates utility on a downstream detection task.}
\section{Methodology}
\label{sec:method}

In this section, we describe PolypSteer through three stages: the backbone and intervention point (Sec.~\ref{sec:method_layers}), Pathology Vector Estimation (Sec.~\ref{sec:method_vectors}), and Spatially Selective Pathology Steering (SSPS) (Sec.~\ref{sec:method_steering}).

\begin{figure}[t]
\centering
\includegraphics[width=\textwidth]{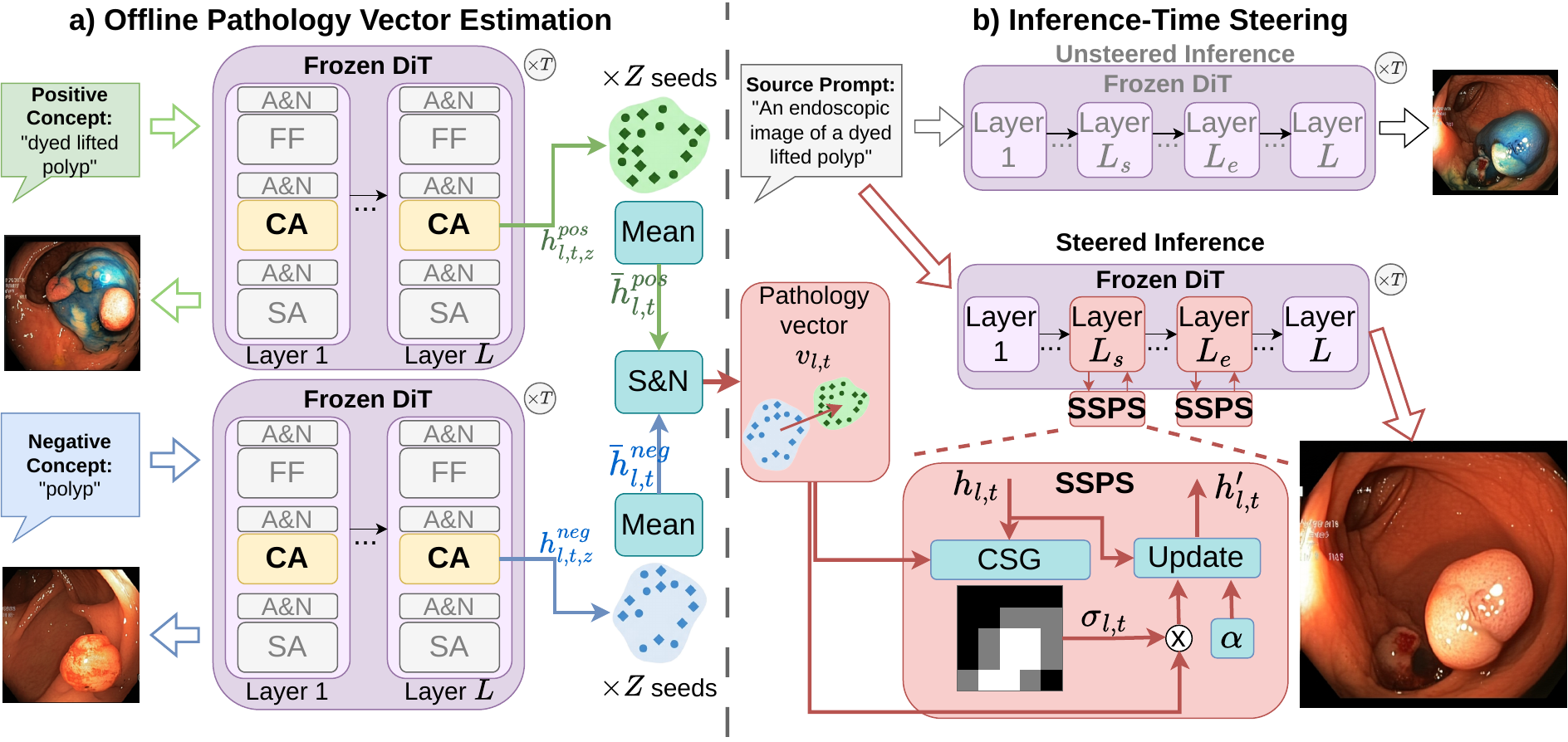}
\caption{PolypSteer method pipeline. \textbf{A\&N}: Add \& Norm. \textbf{FF}: Feed Forward. \textbf{CA}: Cross-Attention. \textbf{SA}: Self-Attention. \textbf{(a) Offline Pathology Vector Estimation:} CA features $h_{l,t,z}$ are collected from the frozen DiT for positive and negative prompts across $Z$ random seeds. A \textit{Mean} step yields $\bar{h}^{pos}_{l,t}$ and $\bar{h}^{neg}_{l,t}$. a \textit{S}\&\textit{N} (Subtract \& Normalize) step then produces the unit pathology vector $v_{l,t}$ (dyed lifted polyp $\to$ polyp). \textbf{(b) Inference-Time Steering:} \textit{Unsteered Inference} runs the frozen DiT unmodified. \textit{Steered Inference} applies Spatially Selective Pathology Steering (SSPS) at layers $l \in \{L_s,\dots,L_e\}$ across all $T$ denoising steps. Inside SSPS, a \textit{CSG} (Cosine-similarity gate) produces the per-token score $\sigma_{l,t}$, which is scaled by $\alpha$ and fed to an \textit{Update} step that subtracts the aligned component from $h_{l,t}$, yielding the counterfactual activation $h'_{l,t}$. While both inference branches are shown together here, only the desired branch is executed in practice.}
\label{fig:method_pipeline}
\end{figure}

\subsection{Backbone and Intervention Point}
\label{sec:method_layers}

PolypSteer builds on PixArt-$\alpha$~\cite{chen2024pixartalpha}, a Diffusion Transformer (DiT)~\cite{DBLP:conf/iccv/PeeblesX23} with $N{=}28$ transformer layers. Each layer contributes self-attention, cross-attention (CA), and feed-forward terms to the residual stream.
The CA output is the sole conduit through which textual semantics reach image tokens, making it the natural intervention point. A direction estimated from contrastive text prompts can be subtracted there to suppress a clinical concept without touching anatomy encoded elsewhere.
We therefore intervene on CA outputs at layers $l \in \{L_s, \dots, L_e\}$ across all denoising timesteps, with the layer range determined empirically in Sec.~\ref{sec:ablation}.

\subsection{Pathology Vector Estimation}
\label{sec:method_vectors}

\noindent\textbf{Contrastive prompt design.}
A clinical concept $P$ is a pathological class label (e.g., Polyp, Normal Cecum, Ulcerative Colitis, Normal Z-line, Esophagitis, Dyed Lifted Polyp).
For each $P$, we assemble $Z$ positive/negative prompt pairs $\{(T^{(z)}_{pos}, T^{(z)}_{neg})\}$ that differ only in the presence of $P$ (e.g., ``\texttt{[context] of a dyed lifted polyp}'' vs.\ ``\texttt{[context] of a polyp}''), with context phrasing (e.g.,\ ``An endoscopic image'') and random seed varied across pairs to marginalise out texture and viewpoint variation.

\noindent\textbf{Pathology vector.}
For each prompt pair, we perform $Z$ forward passes through the frozen model.
\textit{Mean}: the CA features $h_{l,t,n}$ are averaged over all $Z$ seeds and all spatial tokens, yielding $\bar{h}^{pos}_{l,t}, \bar{h}^{neg}_{l,t} \in \mathbb{R}^{d_l}$ per layer $l$ and timestep $t \in \{1,\dots,T\}$.
\textit{Subtract \& Normalize}: following the mean-difference principle~\cite{axbench}, the L2-normalised contrastive difference gives the pathology vector $v_{l,t} = \frac{\bar{h}^{pos}_{l,t} - \bar{h}^{neg}_{l,t}}{\|\bar{h}^{pos}_{l,t} - \bar{h}^{neg}_{l,t}\|_2}$.
This yields a unit vector per layer and timestep that captures the shared semantic difference attributable to $P$. The per-timestep parameterization is necessary because the diffusion process encodes concepts differently across denoising steps, with early timesteps establishing global structure and later steps refining fine-grained details. Since $v_{l,t}$ is concept-specific rather than image-specific, it is computed once offline and reused across all inference calls at no additional cost.

\subsection{Spatially Selective Pathology Steering (SSPS)}
\label{sec:method_steering}

\noindent\textbf{Counterfactual setup.}
Unlike inversion-based methods, PolypSteer requires no source image, no mask annotation, and no fine-tuning: both images are generated from scratch using the same noise seed $z$.
Both branches share the same prompt $T_{pos}$.
While \textit{Unsteered Inference} runs the frozen DiT unmodified, \textit{Steered Inference} inserts SSPS at layers $l \in \{L_s,\dots,L_e\}$ at every denoising timestep to suppress the target pathology. Because both branches share the same seed $z$, any structural difference between the paired outputs arises solely from the SSPS intervention, making the comparison a true minimal-edit counterfactual.

\noindent\textbf{Cosine-similarity gate (CSG).}
Because pathological changes are spatially focal, a global scalar subtraction would over-suppress non-target tokens.
We score each token's alignment with the pathology concept via
$\cos(h_{l,t},\,v_{l,t}) = \langle h_{l,t},\,v_{l,t}\rangle / \|h_{l,t}\|_2$, where $h_{l,t}$ denotes the activation of any visual token at layer $l$, timestep $t$.
We deliberately drop the $1/\|h_{l,t}\|_2$ normalizer, arriving at the dot product $\langle h_{l,t}, v_{l,t} \rangle$, so that tokens expressing the concept more strongly receive proportionally stronger steering:
$\sigma_{l,t} = \max\!\bigl(\langle h_{l,t},\, v_{l,t} \rangle,\; 0\bigr)$.
The $\max(\cdot, 0)$ operator ensures that only tokens positively aligned with $v_{l,t}$ are modified, and tokens with zero or negative alignment pass through unchanged.
We then update each token's activation by subtracting the pathology-aligned component scaled by $\alpha$ (the \textit{Update} step in Fig.~\ref{fig:method_pipeline}), leaving all orthogonal components (anatomy, texture, viewpoint) intact:
$h'_{l,t} = h_{l,t} - \alpha\,\sigma_{l,t}\, v_{l,t}$,
where $\alpha$ is the steering strength.
\section{Experiments}
\label{sec:experiments}

\subsection{Experimental Setup}
\label{sec:setup}

\noindent\textbf{Datasets and model.}
We use Kvasir v3~\cite{pogorelov2017kvasir} (8,000 gastrointestinal endoscopy images, eight classes) for generative training, pathology vector construction, and all generative evaluations. HyperKvasir~\cite{borgli2020hyperkvasir} provides the held-out test set for downstream polyp detection. The generative backbone is PixArt-$\alpha$~\cite{chen2024pixartalpha} fine-tuned with LoRA (rank $r{=}64$) on Kvasir.

\noindent\textbf{Pathology vector construction.}
For each concept pair $(T_{pos}, T_{neg})$ we generate $N{=}50$ images per prompt and apply the method in Sec.~\ref{sec:method} across layers $l \in \{8,\dots,16\}$ and all timesteps $t$ to obtain $v_{l,t}$ for each setting.

\noindent\textbf{Baselines and metrics.}
We compare \textbf{Re-prompting} (i.e. generate another image using a different prompt with the same random seed), and two DDIM-inversion-based editing methods: \textbf{Plug-and-Play (PnP)}~\cite{tumanyan2023plug} and \textbf{h-Edit}~\cite{hedit2024} (RadEdit~\cite{radedit2024} is excluded as it requires per-image mask annotations). A pre-trained oracle classifier (area under the ROC curve, AUC $=0.98$) measures \emph{concept flip rate} (fraction of outputs no longer classified as the source class) and \emph{confidence shift} $\Delta p$ (mean drop in source-class probability). Background preservation is assessed via Bg-LPIPS~\cite{zhang2018perceptual}, Bg-SSIM, and Bg-PSNR (each restricted to the background, i.e., non-lesion, region isolated by UNet++~\cite{zhou2019unetplusplus} (Dice $=0.912$)) for counterfactual experiments. Dye disentanglement additionally reports results under Segformer~\cite{xie2021segformer} (Dice $=0.921$).

\begin{table}[b]
\centering
\caption{Downstream polyp detection on HyperKvasir ($N{=}1{,}000$ synthetic images per condition). CF: counterfactual generation.}
\label{tab:downstream}
\footnotesize
\begin{tabular}{lc cc cc}
\toprule
& & \multicolumn{2}{c}{\textbf{ConvNeXt}} & \multicolumn{2}{c}{\textbf{ViT}} \\
\cmidrule(lr){3-4}\cmidrule(lr){5-6}
\textbf{Methods} & \textbf{CF?} & \textbf{F1}~$\uparrow$ & \textbf{AUC}~$\uparrow$ & \textbf{F1}~$\uparrow$ & \textbf{AUC}~$\uparrow$ \\
\midrule
Real-only                        & \xmark & $0.8977_{\pm.019}$ & $0.8469_{\pm.023}$ & $0.9062_{\pm.022}$ & $0.8942_{\pm.026}$ \\
Re-prompting                     & \xmark & $0.9038_{\pm.016}$ & $0.8714_{\pm.020}$ & $0.9147_{\pm.018}$ & $0.9083_{\pm.022}$ \\
PnP               & \cmark & $0.9174_{\pm.014}$ & $0.9086_{\pm.017}$ & $0.9437_{\pm.013}$ & $0.9518_{\pm.017}$ \\
h-Edit            & \cmark & $0.9097_{\pm.015}$ & $0.8902_{\pm.019}$ & $0.9295_{\pm.016}$ & $0.9312_{\pm.020}$ \\
\textbf{PolypSteer} & \cmark & $\mathbf{0.9263}_{\pm.012}$ & $\mathbf{0.9341}_{\pm.015}$ & $\mathbf{0.9591}_{\pm.011}$ & $\mathbf{0.9755}_{\pm.013}$ \\
\bottomrule
\end{tabular}
\end{table}

\subsection{Downstream Polyp Detection}
\label{sec:exp_downstream}

In the first experiment, we assess whether PolypSteer-generated augmentations improve downstream polyp detection.
We train binary polyp detectors (ConvNeXt~\cite{liu2022convnet} and ViT~\cite{dosovitskiy2020image}) under five augmentation methods: \textbf{Real-only} (20 images/class), \textbf{Re-prompting}, \textbf{PnP}, \textbf{h-Edit}, and \textbf{PolypSteer}. Each augmention method adds $N{=}1{,}000$ synthetic images so that differences reflect augmentation \emph{structure}, not \emph{quantity}. Since HyperKvasir is an extension of Kvasir, we remove all exact and near-duplicate matches from the test set before evaluation to ensure a strictly out-of-distribution assessment. Table~\ref{tab:downstream} shows that PolypSteer achieves the best AUC under both backbones. This proves that paired supervision forces detectors to learn useful pathology features that generalize well into OOD setting.

\begin{figure}[t]
\centering
\includegraphics[width=\textwidth]{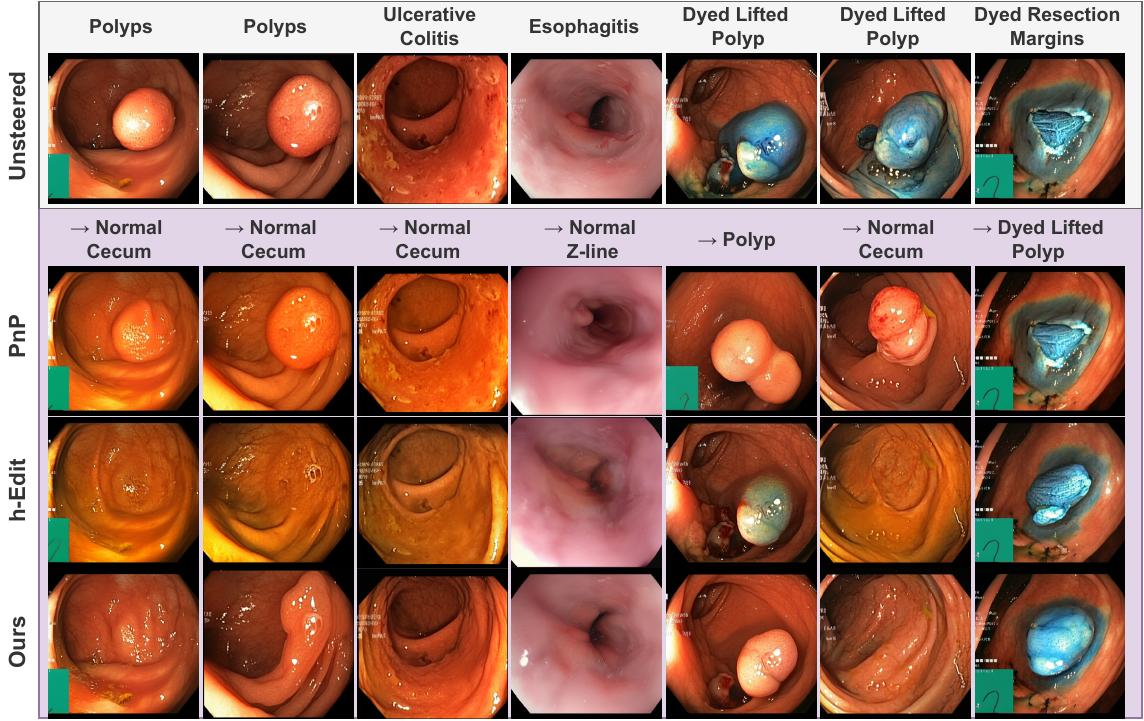}
\caption{Qualitative comparison across concept pairs. Rows: Unsteered, PnP, h-Edit, and PolypSteer (Ours). Columns show Polyp $\to$ Normal Cecum, Ulcerative Colitis $\to$ Normal Cecum, Esophagitis $\to$ Normal Z-line, and dye steerings.}
\label{fig:counterfactual}
\end{figure}
\begin{table}[t]
\centering
\caption{Counterfactual generation (UC: Ulcerative Colitis, Esoph.: Esophagitis). Bg metrics are not applicable for pairs~(B) and~(C): the polyp segmentor cannot produce valid masks for diffuse mucosal inflammation or esophageal anatomy. Re-prompting is excluded because it rerolls the trajectory and produces different anatomy.}
\label{tab:counterfactual}
\resizebox{\linewidth}{!}{%
\begin{tabular}{lcccccccccc}
\toprule
& \multicolumn{5}{c}{\textbf{(A) Polyp $\to$ Norm.\ Cecum}}
& \multicolumn{2}{c}{\textbf{\shortstack{(B) UC $\to$\\Normal Cecum}}}
& \multicolumn{2}{c}{\textbf{\shortstack{(C) Esoph. $\to$\\Normal Z-line}}} \\
\cmidrule(lr){2-6}\cmidrule(lr){7-8}\cmidrule(lr){9-10}
\textbf{Method}
  & \textbf{Flip}~$\uparrow$ & \textbf{$\Delta p$}~$\uparrow$
  & \textbf{Bg-LPIPS}~$\downarrow$ & \textbf{Bg-SSIM}~$\uparrow$ & \textbf{Bg-PSNR}~$\uparrow$
  & \textbf{Flip}~$\uparrow$ & \textbf{$\Delta p$}~$\uparrow$
  & \textbf{Flip}~$\uparrow$ & \textbf{$\Delta p$}~$\uparrow$ \\
\midrule
PnP               & 0.627 & 0.543 & 0.1518 & 0.8506 & 19.32 & 0.682 & 0.617 & 0.718 & 0.651 \\
h-Edit            & 0.714 & 0.638 & 0.1565 & 0.8222 & 19.51 & 0.741 & 0.663 & 0.796 & 0.712 \\
\textbf{PolypSteer} &  \textbf{0.800} &  \textbf{0.721} & \textbf{0.1449} & \textbf{0.8951} & \textbf{22.95}
                  & \textbf{0.925} & \textbf{0.902} & \textbf{0.950} & \textbf{0.928} \\
\bottomrule
\end{tabular}%
}
\end{table}
\begin{table}[t]
\centering
\caption{Dye disentanglement under two background segmentors. DDR = dye detection rate (lower = dye removed). Eff.~LPIPS $= \text{LPIPS}/(1{-}\text{DDR})$ ($\downarrow$). Eff.~SSIM $= (1{-}\text{DDR}){\times}\text{SSIM}$, Eff.~PSNR $= (1{-}\text{DDR}){\times}\text{PSNR}$ ($\uparrow$).}
\label{tab:dye}
\resizebox{\columnwidth}{!}{%
\begin{tabular}{l ccc c ccc}
\toprule
& \multicolumn{3}{c}{\textbf{UNet++}} & & \multicolumn{3}{c}{\textbf{Segformer}} \\
\cmidrule(lr){2-4}\cmidrule(lr){6-8}
\textbf{Method} & \textbf{Eff.\ LPIPS}~$\downarrow$ & \textbf{Eff.\ SSIM}~$\uparrow$ & \textbf{Eff.\ PSNR}~$\uparrow$ & \textbf{DDR}~$\downarrow$ & \textbf{Eff.\ LPIPS}~$\downarrow$ & \textbf{Eff.\ SSIM}~$\uparrow$ & \textbf{Eff.\ PSNR}~$\uparrow$ \\
\midrule
h-Edit            & 0.4540 & 0.0957 & 2.78  & 0.900 & 0.5190 & 0.0952 & 2.73 \\
PnP               & 0.3021 & 0.1872 & 4.93  & 0.800 & 0.3235 & 0.1865 & 4.82 \\
\midrule
\textbf{PolypSteer} & \textbf{0.2027} & \textbf{0.6584} & \textbf{15.57} & \textbf{0.250} & \textbf{0.2193} & \textbf{0.6521} & \textbf{15.18} \\
\bottomrule
\end{tabular}%
}
\end{table}

\begin{table}[t]
\centering
\caption{Ablation studies (Polyp $\to$ Normal Cecum). \textbf{(a)}~$L_s$--$L_e$ ablation ($\alpha{=}2.5$, $N{=}50$), $\dagger$W = window width, i.e.\ number of layers steered. \textbf{(b)}~Steering strength (layers 8--16, $N{=}50$). \textbf{(c)}~Number of seeds (layers 8--16, $\alpha{=}2.5$).}%
\label{tab:ablation_layers}%
\label{tab:ablation_alpha}%
\label{tab:ablation_seeds}%
\resizebox{\linewidth}{!}{%
\begin{tabular}[t]{lccc}
\toprule
\multicolumn{4}{l}{\textbf{(a) $L_s$--$L_e$ ablation}} \\
\midrule
\textbf{Layers} & \textbf{W\textsuperscript{$\dagger$}} & \textbf{Flip}~$\uparrow$ & \textbf{$\Delta p$}~$\uparrow$ \\
\midrule
0--8   & 8  & 0.025 & 0.018 \\
\textbf{8--16}  & \textbf{8}  & \textbf{0.800} & \textbf{0.721} \\
12--20 & 8  & 0.013 & 0.009 \\
\midrule
0--12  & 12 & 0.377 & 0.396 \\
8--20  & 12 & 0.700 & 0.623 \\
\bottomrule
\end{tabular}%
\hspace{2em}%
\begin{tabular}[t]{lcc}
\toprule
\multicolumn{3}{l}{\textbf{(b) Steering strength}} \\
\midrule
\textbf{$\alpha$} & \textbf{Flip}~$\uparrow$ & \textbf{$\Delta p$}~$\uparrow$ \\
\midrule
0.5 & 0.012 & 0.008 \\
1.0 & 0.038 & 0.021 \\
2.0 & 0.513 & 0.516 \\
\textbf{2.5} & \textbf{0.800} & \textbf{0.721} \\
3.0 & 0.700 & 0.625 \\
\bottomrule
\end{tabular}%
\hspace{2em}%
\begin{tabular}[t]{lcc}
\toprule
\multicolumn{3}{l}{\textbf{(c) Number of seeds}} \\
\midrule
\textbf{$N$} & \textbf{Flip}~$\uparrow$ & \textbf{$\Delta p$}~$\uparrow$ \\
\midrule
30           & 0.762 & 0.689 \\
\textbf{50}  & \textbf{0.800} & \textbf{0.721} \\
80           & 0.793 & 0.715 \\
100          & 0.791 & 0.712 \\
\bottomrule
\end{tabular}%
}
\end{table}
\begin{figure}[t]
\centering
\includegraphics[width=\columnwidth]{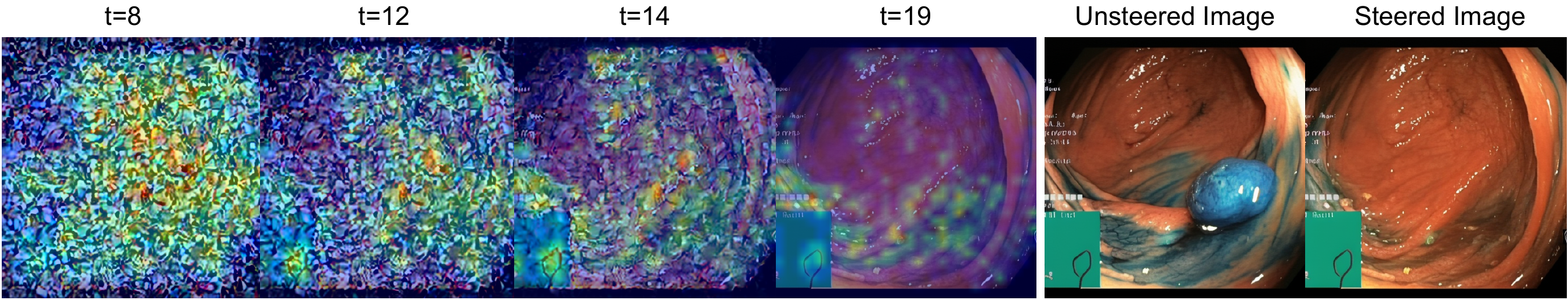}
\caption{\textit{Left}: Per-token cosine similarity maps $\sigma_{8,t}$ at layer~8 for selected diffusion steps $t\in\{8,12,14,19\}$ (left to right). Warmer colours indicate stronger alignment with the pathology vector. \textit{Right}: corresponding unsteered (dyed lifted polyps) and steered (normal cecum) endoscopy images.}
\label{fig:attention}
\end{figure}
\subsection{Counterfactual Generation Across Clinical Concepts}
\label{sec:exp_counterfactual}

In the second experiment, we evaluate concept steering quality across multiple clinical concept pairs, measuring concept flip rate and background preservation.
We evaluate three concept pairs: (A)~Polyp~$\leftrightarrow$~Normal Cecum, (B)~Ulcerative Colitis~$\leftrightarrow$~Normal Cecum, and (C)~Esophagitis~$\leftrightarrow$~Normal Z-line ($N{=}1{,}000$ images, $\alpha{=}{\pm}2.5$). Background metrics are reported for pair~(A) only. 

PolypSteer leads on pair~(A): Flip $=0.800$, $\Delta p{=}0.721$, Bg-LPIPS $=0.1449$, and best Bg-SSIM and Bg-PSNR. Inversion-based methods accumulate DDIM reconstruction error, degrading both concept-change rate and background fidelity. Pairs~(B) and~(C) yield flip rates of 0.925 and 0.950, confirming generalisation across clinical concepts and anatomical sites (Table~\ref{tab:counterfactual} and Fig.~\ref{fig:counterfactual}).

\subsection{Dye Disentanglement}
\label{sec:exp_dye}

In the third experiment, we test disentanglement of co-occurring attributes (polyp morphology and dye staining) when text-based disentanglement is impossible.
The ``Dyed Lifted Polyp'' class encodes two co-occurring attributes (polyp morphology and indigo carmine staining) with no ``undyed'' counterpart in the training distribution, making text-based disentanglement impossible. We construct a dye pathology vector from $T_{pos}{=}$``dyed lifted polyps'' and $T_{neg}{=}$``polyps''. Because the pair differs only in dye presence, the resulting vector is orthogonal to polyp morphology by construction, suppressing only the dye attribute while preserving polyp structure. PolypSteer achieves a dye detection rate (DDR, fraction of outputs still classified as dyed) of 0.250, substantially outperforming h-Edit (DDR $=0.900$) and PnP (DDR $=0.800$) (Table~\ref{tab:dye} and Fig.~\ref{fig:counterfactual}). PolypSteer also natively supports compositional edits (e.g., $\text{Dyed Lifted Polyp} \to \text{Normal Cecum}$) with strict anatomical consistency.

\subsection{Ablation Studies}
\label{sec:ablation}

Table~\ref{tab:ablation_layers} sweeps layer windows at fixed $\alpha{=}2.5$, showing that layers 8--16 are the semantic formation zone. Windows entirely outside this range yield near-zero flip, and a width-12 window at 0--12 achieves only 37.7\% flip vs.\ 70.0\% for 8--20, confirming a position effect. Table~\ref{tab:ablation_alpha} shows flip rises sharply from $\alpha{=}2.0$ (51.3\%) to $\alpha{=}2.5$ (80.0\%) and degrades at $\alpha{=}3.0$ (70.0\%). The pathology vector stabilises after ${\sim}50$ seeds (Table~\ref{tab:ablation_seeds}). Per-token cosine similarity scores $\sigma_{l,t}$ provide a built-in spatial explanation, with the activation footprint being broad at early diffusion steps and contracting to sparse patches by the final steps.

\subsection{Interpretability}
\label{sec:interpretability}

The per-token cosine similarity score $\sigma_{l,t}$ can be rendered directly as a spatial map, yielding a built-in visualisation of \emph{where} the model steers at each diffusion step. As Fig.~\ref{fig:attention} shows, the activation footprint covers nearly the entire image at early steps and becomes progressively sparser toward the final steps, reflecting that fewer tokens require adjustment once the concept has been established. Inversion-based baselines have no analogous interpretable gate.

\section{Conclusion}
\label{sec:conclusion}

We presented \textbf{PolypSteer}, a training-free steering framework for medical image synthesis.
It extracts pathology vectors from contrastive prompt pairs in the cross-attention space of a frozen diffusion transformer, then steers activations via cosine similarity to produce counterfactuals, with no source images, DDIM inversion, mask annotations, or retraining.
On Kvasir v3 and HyperKvasir, PolypSteer outperforms baselines on downstream polyp detection, counterfactual generation, and dye disentanglement.
Additionally, per-token cosine similarity provides built-in spatial interpretability at each diffusion step.
Future work will extend to 3D volumetric data, video endoscopy sequences, and cross-institutional deployment.

\clearpage
\bibliographystyle{splncs04}
\bibliography{references}

\end{document}